# COMPRESSION-AWARE PROJECTION WITH GREEDY DIMENSION REDUCTION FOR CONVOLUTIONAL NEURAL NETWORK ACTIVATIONS

*Yu-Shan Tai, Chieh-Fang Teng, Cheng-Yang Chang, and An-Yeu (Andy) Wu*

Graduate Institute of Electrical Engineering, National Taiwan University, Taipei, Taiwan

{clover, jeff, kevin}@access.ee.ntu.edu.tw, andywu@ntu.edu.tw

## ABSTRACT

Convolutional neural networks (CNNs) achieve remarkable performance in a wide range of fields. However, intensive memory access of activations introduces considerable energy consumption, impeding deployment of CNNs on resource-constrained edge devices. Existing works in activation compression propose to transform feature maps for higher compressibility, thus enabling dimension reduction. Nevertheless, in the case of aggressive dimension reduction, these methods lead to severe accuracy drop. To improve the trade-off between classification accuracy and compression ratio, we propose a compression-aware projection system, which employs a learnable projection to compensate for the reconstruction loss. In addition, a greedy selection metric is introduced to optimize the layer-wise compression ratio allocation by considering both accuracy and #bits reduction simultaneously. Our test results show that the proposed methods effectively reduce 2.91×~5.97× memory access with negligible accuracy drop on MobileNetV2/ResNet18/VGG16.

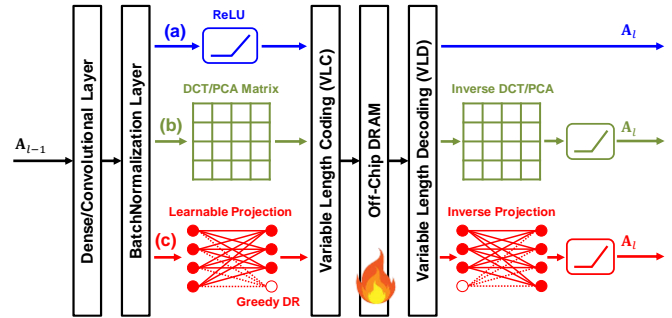

**Fig. 1.** Overview of different activation compression (AC) methods: (a) Lossless encoder AC; (b) Transform-based AC; (c) The proposed compression-aware projection with greedy dimension reduction.

## 1. INTRODUCTION

In recent years, CNNs have achieved amazing performance in various applications, such as face recognition [1], image classification [2], disease detection [3], and so on. Despite of the superior performance, extensive computation requirements and intermediate data communication between deep learning accelerator (DLA) and off-chip memory hinder CNNs from being embedded on edge devices. Therefore, model compression techniques, e.g., pruning and quantization, have been intensively researched and widely used.

As presented in [4][5], data movement of activations consumes almost 70% of the total energy footprint. Consequently, as another line of work orthogonal to model compression, activation compression (AC) is receiving increasing attention. To remove redundant elements without losing essential information, prior works [6]-[10] have successfully exploited the sparsity induced by the rectified linear unit (ReLU) function, as shown in Fig. 1(a). Nevertheless, the sparsity of activations is dynamic and is highly dependent on the input data [6]. Without satisfactory sparsity, a lossless encoder cannot guarantee to reach the expected compression ratio.

To enhance activation sparsity, transform-based AC leverages discrete cosine transform (DCT) [7][11] or principal component analysis (PCA) [12]-[14] before sending activations to variable length coding (VLC), as shown in Fig. 1(b). After transforming activations to another domain, important/unimportant parts of activations become separable and thus improve compressibility.

Although existing transform-based methods have been proven to be effective for AC, there are still some challenges:
1) *Sacrificed accuracy for dimension reduction:* PCA or DCT-based transformation matrix does not consider the following compression. Therefore, further dimension reduction (DR) results in severe performance degradation.
2) *Threshold-based dimension reduction*: Existing methods determine the DR ratio of each layer according to the same accumulated eigenvalue threshold [12][13]. However, they neglect the layer-wise differences among eigenvalue distributions and sizes of activations vary among different CNN layers. Ignoring the information mentioned above makes the compressing process sub-optimal.

To address the above issues, we propose a compression-aware projection system with greedy dimension reduction (DR) as shown in Fig. 1(c). Our main contributions are:
1) *Learnable projection*: We use a learnable projection to compensate for the reconstruction loss induced by compression. Our experiment results show that learnable projection can improve the memory access of MobileNetV2/ResNet18/VGG16 by 2.85×~5.06×.
2) *Greedy dimension reduction (DR)*: We design a selection metric specialized for greedy DR, which iteratively utilizes DR to reduce the storage overhead of the layer with the lowest metric in each round. Combining greedy DR with our learnable projection leads to a better trade-off between classification accuracy and compression ratio, further improving the memory reduction rate to 2.91×~5.97×.

The rest of this paper is organized as follows. Section **2** briefly introduces the existing activation compression methods. Section **3** illustrates the proposed compress-aware projection, which comprises a learnable projection and greedy DR based on the selection metric. The experiments and analyses are illustrated in Section **4**. Finally, Section **5** concludes our work.

This research work is financially supported in part by Novatek, under grant 110HT945009, and in part by Ministry of Science and Technology, Taiwan, under grant MOST 110-2218-E-002-034-MBK.

## 2. RELATED WORK

### 2.1. Lossless Encoder for Activation Compression

Activation, or feature map, is the intermediate output of each layer. Since the ReLU function makes activations sparse, several researchers exploited this characteristic to develop sparsity-based lossless encoder, as shown in Fig. 1(a).

One of the most popularly used lossless encoders is run-length encoding (RLE), which keeps the length of zero interval instead of storing consecutive zeros. In [6], RLE is used to compress both weight and activation. Another well-known lossless encoder is zero value compression (ZVC) [10], which utilizes a non-zero-mask to indicate the location of non-zero values and is helpful to compress data with randomly distributed sparsity [7]. Huffman encoding is also a widely known compression method [8][9]. This encoding is based on the frequency of elements and achieves a higher compression ratio with biased data.

However, the sparsity of activations is dynamic, which depends on model architecture and input data characteristics. Thus, the above methods are highly sensitive to sparsity and may perform poorly with dense activations.

### 2.2. Transform-based Activation Compression

To overcome the restriction of sparsity pattern and enhance compression ratio, transform-based methods are introduced, as shown in Fig. 1(b). Since activation is generated by mapping relatively small-sized inputs to high dimensions, there is high redundancy among channels [13]. Rather than sending activation to sparsity-based encoder directly, transformed-based methods adopt domain transformation first to separate important/unimportant components. Therefore, the compression ratio can be further improved by removing the latter.

In [7], the authors operated DCT to project activation to the frequency domain. To further fold the transformation matrix into convolution layer, [11] implemented 1D-DCT on the channel domain rather than the spatial domain and utilized a mask to facilitate channel reordering. However, the mask is hard to design and greatly impacts performance.

Instead of using DCT as the transformation matrix, some other works replaced it with PCA. In [12], the authors adopted a pre-computed PCA matrix to transform activations. Let $\mathbf{A}$ stand for the activation, whose size is $n \times d \times w \times h$ and $n, d, w, h$ represent the batch size, channel number, width, and height, respectively. Since PCA operates on the channel domain, $\mathbf{A}$ would need to be reshaped to $\mathbf{A}_c \in d \times (n \times w \times h)$ first. Then, the corresponding transformation matrix $\mathbf{U}$ can be obtained by PCA:

$$\mathbf{U}, \mathbf{\Sigma} = PCA(\mathbf{A}_c), \quad (1)$$

where $\mathbf{U}$ stands for the orthogonal basis and $\mathbf{\Sigma} = \{\sigma_1^2, \sigma_2^2, ..., \sigma_d^2\}$ are its corresponding eigenvalues. By multiplying $\mathbf{A}_c$ with $\mathbf{U}$, we can obtain the transformed activations $\mathbf{A_U}$:

$$\mathbf{A_U} = \mathbf{U} \times \mathbf{A}_c, \quad (2)$$

Afterward, $\mathbf{A_U}$ would undergo quantization and variable length coding (VLC) as presented in Fig. 1(b). Finally, the #bits of $\mathbf{A}$ needed to send to off-chip memory can be displayed by:

$$B(\mathbf{A_U}) = \#bits\left(VLC(Q(\mathbf{A_U}))\right). \quad (3)$$

If needed to fetch, activations would be reconstructed by the symmetric inverse process. Since the PCA matrix is data-dependent, it is more likely to reach higher compressibility than DCT, whose matrix is fixed and suitable for data with locality. Moreover, the importance of each channel could be verified by comparing its eigenvalue, where larger $\sigma$ implies containing a higher amount of information. Thus, threshold-based dimension reduction utilizes this concept to keep minimum #channels until cumulated eigenvalue reaches the defined percentage $T$, which can be specified as:

$$d'_l = \frac{\arg\min_k \sum_{i=1}^{k} \sigma_i^2}{\sum_{i=1}^{d_l} \sigma_i^2} \geq T, \quad l \in \{1, 2, ... L\}, \quad (4)$$

where $d_l$ and $d'_l$ denote #channels of layer $l$ before and after dimension reduction, and $L$ stands for the total number of layers. By removing these unimportant channels, the corresponding dimension of activations could be further reduced to a smaller scale, which directly decreases memory access requirement.

Although PCA transformation can address the lack of sparsity and locality, there still exists some room for improvement. First, since the PCA matrix is obtained just by analyzing input data distribution, DR and quantization are not considered. This leads to dramatical accuracy loss after further compression. Secondly, reducing the dimension of each layer as Eq. (4) is not ideal. Since each layer owns different eigenvalue distribution, the impact of DR also differs. On the other hand, the sizes of feature maps have a significant influence on the compression strength. Ignoring the difference among layers would not only make the compression process inefficient but also hurt accuracy.

## 3. PROPOSED LEARNABLE PROJECTION WITH GREEDY BASED DIMENSION REDUCTION

### 3.1. Learnable Projection

To maintain accuracy with a high compression ratio, we propose a compression-aware projection system to compensate for the loss, which is shown in Fig. 1(c). The main idea of our method is to make the transformation matrix trainable while keeping other model weights frozen. The detail of the system is illustrated at the bottom of Fig. 2, where $\mathbf{A}_l$, $\mathbf{A}'_l$ denote the activations of the original model and that of the learnable projection, $\mathbf{P}_l$ and $\mathbf{P}_l^{inv}$ are the learnable projection matrix and inverse matrix of layer $l$. $*$ stands for convolution followed by batch normalization (BN), and $Q$ as well as $Q^{-1}$ represent uniform quantization and its inverse operation, respectively.

The details of training steps are specified as follows. First, we initialize $\mathbf{P}_l$ and $\mathbf{P}_l^{inv}$ with PCA transformation matrix and its transpose matrix to make our training start at a good point. Next, we train the two transform matrices with hint loss and knowledge distillation (KD) loss, which can be written as:

$$\mathcal{L} = \mathcal{L}_{hint} + \mathcal{L}_{KD}, \quad (5)$$

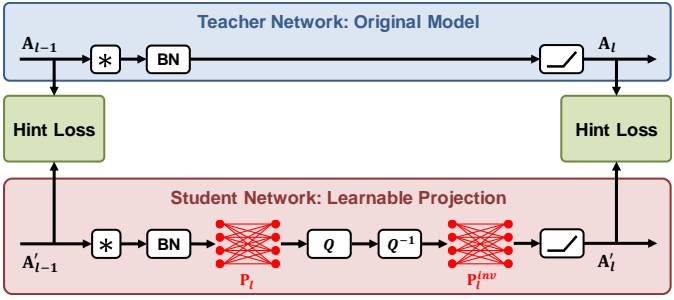

**Fig. 2.** The relation of the original model and learnable projection.

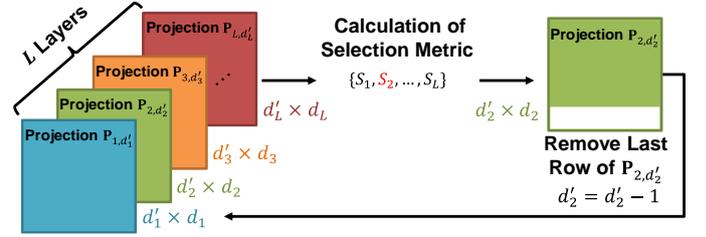

**Fig. 3.** Selection metric for greedy dimension reduction.

$$\mathcal{L}_{hint} = \sum_{l=1}^{L} \|\mathbf{A}_l - \mathbf{A}'_l\|_2, \quad (6)$$

$$\mathcal{L}_{KD} = D_{KL}(\mathbf{o}\|\mathbf{o}'), \quad (7)$$

where $D_{KL}(\cdot \| \cdot)$ is the Kullback-Leibler divergence (KL-divergence) and $\mathbf{o}$ as well as $\mathbf{o}'$ stand for the output vector of the original model and that of the learnable projection model. We calculate hint loss as specified in [14], which uses activations after ReLU to take nonlinear function into consideration. As for the KD loss, we analyzed cross-entropy loss and KL divergence to measure the bias of output distribution and found the latter led to better results. Therefore, instead of training with labeled data, our proposed learnable projection only requires the activations and soft output generated by the original model. Labeled data are hard to access and sometimes involve privacy issues, so our training strategy is feasible to real-world scenarios.

The PCA matrix for each layer is pre-computed offline, and the transformation matrix can also be folded into the convolution and BN operations [12]. Therefore, the only computational overhead introduced by learnable projection is the inverse process (i.e., $\mathbf{P}_l^{inv}$). In Sec. **4**, we would further elaborate on the analysis of additional computation overhead.

### 3.2. Selection Metric for Greedy Dimension Reduction

Though threshold-based dimension reduction can remove unimportant channels, this strategy leads to severe accuracy drop under a high compression ratio. To tackle this issue, we design a greedy selection metric to consider both accuracy drop and memory reduction to achieve a better trade-off.

Our strategy is to greedily choose one layer for dimension reduction at each step and iterate until the required #bits are less than the memory constraint. To achieve our goal, we design a selection metric to prioritize which layer for DR:

$$S_l = \Delta accuracy / \Delta N, \quad (8)$$

where $\Delta accuracy$ and $\Delta N$ denote the loss of accuracy and reduced #bits after operating DR on layer $l$. Small $S_l$ implies sacrificing minor accuracy and reducing huge memory requirements. Take Fig. 3 as an example. There are $L$ projection matrices $\{\mathbf{P}_{1,d'_1}, \mathbf{P}_{2,d'_2}, \dots, \mathbf{P}_{L,d'_L}\}$, and each $\mathbf{P}_{l,d'_l}$ contains $d'_l$ rows. By computing the selection metrics among $L$ layers, we obtain $S = \{S_1, S_2, \dots, S_L\}$. If the minimum occurred at layer 2, we would remove the last row of $\mathbf{P}_{2,d'_2}$. Afterward, $d'_2$ is updated to $d'_2 - 1$. We keep greedily selecting layers to reduce dimension until required #bits meets the user-defined constraint.

Since evaluating the accuracy drop in each step is time-consuming for large-scale tasks (e.g., ImageNet), a simple yet effective alternative evaluation metric is significant. As mentioned in [14], the product of the layer-wise cumulated eigenvalue is highly related to final accuracy. Therefore, we utilize the percentage of eigenvalue on layer $l$ to approximate the accuracy drop induced by DR:

$$\Delta accuracy \approx \sigma_{l,d'_l} / \sum_{c=1}^{d'_l} \sigma_{l,c}. \quad (9)$$

As for $\Delta N$, we measure the difference of required #bits after DR to quantify the gain of compression:

$$\Delta N = B\left(\mathbf{P}_{l,d'_l} \times \mathbf{A}'_l\right) - B\left(\mathbf{P}_{l,d'_l-1} \times \mathbf{A}'_l\right). \quad (10)$$

Compared to equation (4), the proposed selection metric for greedy DR reaches a better trade-off since it takes both accuracy and compressibility into consideration. In Sec. **4**, we would compare the performance of our greedy-based method with threshold-based work to evaluate its effectiveness and then analyze the different distribution of DR between them.

## 4. SIMULATION RESULTS

In the following experiments, we implement our proposed method on pre-trained MobileNetV2, ResNet18, and VGG16, whose weights and activations are both quantized to 8 bits. The dataset we use is ImageNet (ILSVRC 2012) [15]. We randomly sample 50,000 out of 1,281,167 training data to fine-tune our projection matrices, and use Huffman encoder as our VLC. For each implementation, we set learning rate as 1e-3 and number of epochs is 3, the batch sizes of MobileNetV2, ResNet18, and VGG16 are set as 150, 250, 64, respectively. We use stochastic gradient descent (SGD) as our optimizer. All the experiments are operated with PyTorch1.9.0 and Python3.9.

### 4.1. Analysis of Compression between Different Methods

In this section, we evaluate the compression performance among different methods as shown in Fig. 4. The performance of Huffman coding and PCA [12], corresponding to Fig. 1(a) and Fig. 1(b), are conducted as baseline for evaluation. In Fig. 4, the number marked at each point indicates DR under different eigenvalue threshold $T$, which is set as [0.97, 0.98, 0.99, 0.995, 1]. Note that there is no need to set the eigenvalue threshold for our greedy DR. Instead, we keep discarding rows until #bits reaches a similar quantity as baseline for a fair comparison.

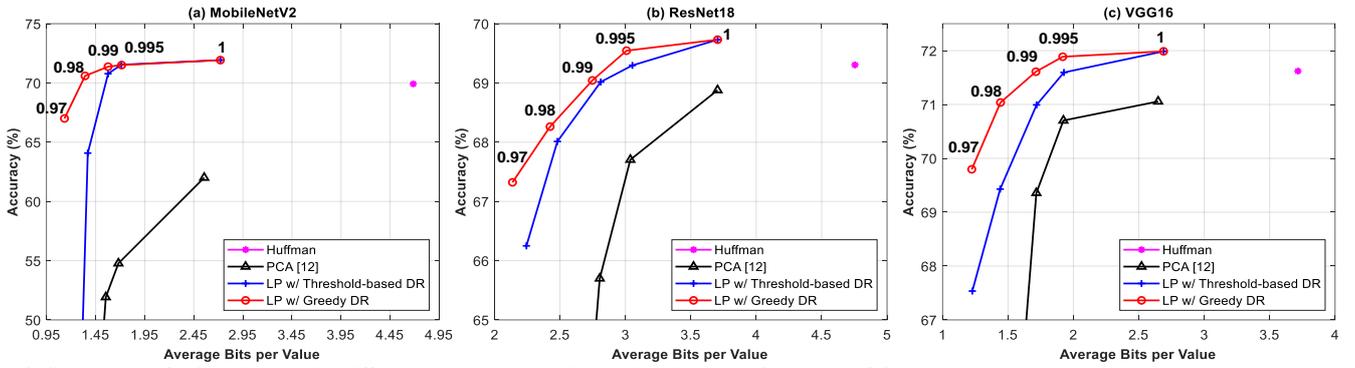
**Fig. 4.** Comparison of AC methods under different models: (a) MobileNetV2, (b) ResNet18, and (c) VGG16.

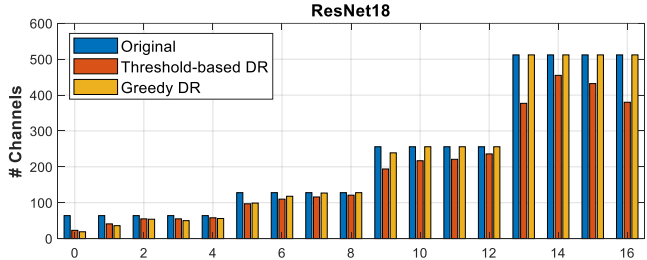

**Fig. 5.** The visualization of channel distribution of greedy DR and threshold-based DR simulated on ResNet18 with $T = 0.995$.

From Fig. 4, we can observe only using Huffman coding receives a low compression ratio. The reason is that Huffman coding is lossless and sensitive to sparsity. On the other hand, PCA enhances compressibility by decorrelation and DR, but accuracy suffers from catastrophic drops as eigenvalue threshold decreases.

By introducing the trainable mechanism, proposed learnable projection (LP) with threshold-based DR can compensate the compression loss effectively and thus preserve higher accuracy than PCA. We reach 0.4%/0.4%/0.6% accuracy drop with average 1.58/2.81/1.72 bits per value, which reduce memory access of original MobileNetV2/ResNet18/VGG16 model by 5.06/2.85/4.65 times, respectively.

Moreover, after equipping learnable projection with greedy DR as presented in Sec. **3.2**, the performance can be further improved. Since considering accuracy drop and bits reduction simultaneously, we can reach a better trade-off than equation (4). In summary, we reach negligible 0.6%/0.4%/0.6% accuracy drop with average 1.34/2.75/1.44 bits per value, which reduce memory access of original MobileNetV2/ResNet18/VGG16 model by 5.97/2.91/5.56 times, respectively.

### 4.2. Visualization of Dimension Reduction Policy

In this part, we would analyze the difference between proposed greedy DR and threshold-based DR by visualizing the #channels among different layers after DR. Due to the limited space, we take ResNet18 with threshold set to 0.995 as an example. In Fig. 5, greedy DR tends to remain a greater #channels for deep layers than threshold-based DR. Since the size of deep layers is smaller than shallow layers, they are more likely to receive a large selection metric. Specifically, the size of the first feature map is $112 \times 112$ while that of the last layer is $7 \times 7$. Consequently, we can conclude that compressing shallow layers leads to higher #bits reduction and makes our greedy DR more effective.

TABLE I. COMPUTATION ANALYSIS UNDER DIFFERENT EIGENVALUE THRESHOLD

| Model | 0.97 | 0.98 | 0.99 | 0.995 | 1 |
|---|---|---|---|---|---|
| MobileNetV2 | 59.7% | 74.3% | 89.5% | 98.0% | 126.8% |
| ResNet18 | 78.5% | 86.4% | 93.8% | 99.3% | 114.3% |
| VGG16 | 57.8% | 70.0% | 83.1% | 92.2% | 114.4% |

### 4.3. Analysis of Additional Computation Overhead

In Sec. **4.1**, we demonstrate the effectiveness of activation compression of our method. To further analyze the additional computation overhead induced by the learnable projection and inverse transformation matrix, we analyze our work under different eigenvalue thresholds $T$. As shown in Table I, each value represents the relative computation over the original model, which can be specified as:

$$\frac{(C_{Original} + C_{Learnable}) + C_{Inverse}}{C_{Original}} \times 100\% \qquad (11)$$

$$= \frac{C_{Folded} + C_{Inverse}}{C_{Original}} \times 100\%, \qquad (12)$$

where $C_{Original}$, $C_{Learnable}$, $C_{Inverse}$, $C_{Folded}$ stand for the computation of original model, learnable projection, inverse projection, and after folding learnable projection into convolution. When the eigenvalue threshold is set to 1, it means no dimension reduction and the overhead is purely induced by inverse projection. However, as threshold decreases, dimension reduction can reduce matrix size and make the computation of our method even less than the original model.

### 5. CONCLUSION

In this paper, we propose a compression-aware projection system. By training a learnable projection, the reconstruction loss induced by compression could be compensated. Moreover, we design a selection metric specialized for greedy DR, taking both accuracy and compressibility into account. Experimental results show that our method reduces 2.91×~5.97× memory access with negligible accuracy drop on MobileNetV2/ResNet18/VGG16.

# 6. REFERENCES

[1] O. M. Parkhi, A. Vedaldi and A. Zisserman, "Deep Face Recognition," in *Proc. of the British Machine Vision Conference (BMVC)*, 2015.

[2] A. Krizhevsky, I. Sutskever and G. E. Hinton, "ImageNet Classification with Deep Convolutional Neural Networks," in *Advances in Neural Information Processing Systems*, 2012.

[3] S.-C. B. Lo et al., "Artificial convolution neural network techniques and applications for lung nodule detection," *IEEE Trans. Med. Imag.*, vol. 14, no. 4, pp. 711–718, Dec. 1995.

[4] T.-J. Yang, Y.-H. Chen, J. Emer and V. Sze, "A method to estimate the energy consumption of deep neural networks," in *Proc. Asilomar Conference on Signals, Systems, and Computers*, 2017.

[5] V. Sze, Y.-H. Chen, T.-J. Yang and J. S. Emer, "Efficient Processing of Deep Neural Networks: A Tutorial and Survey," *Proceedings of the IEEE*, vol. 105, pp. 2295-2329, 2017.

[6] A. Parashar, et al., "SCNN: An accelerator for compressed-sparse convolutional neural networks," in *Proc. ACM/IEEE Annual International Symposium on Computer Architecture (ISCA)*, 2017, pp. 27–40.

[7] R. D. Evans, L. Liu and T. M. Aamodt, "JPEG-ACT: Accelerating Deep Learning via Transform-based Lossy Compression," in *Proc. ACM/IEEE Annual International Symposium on Computer Architecture (ISCA)*, 2020.

[8] M. Chandra, "Data Bandwidth Reduction in Deep Neural Network SoCs using History Buffer and Huffman Coding," in *Proc. International Conference on Computing, Power and Communication Technologies (GUCON)*, 2018.

[9] S. Han, H. Mao and W. J. Dally, "Deep Compression: Compressing Deep Neural Network with Pruning, Trained Quantization and Huffman Coding," in *International Conference on Learning Representations (ICLR)*, San Juan, Puerto Rico, May 2-4, 2016.

[10] M. Rhu, M. O'Connor, N. Chatterjee, J. Pool, Y. Kwon and S. W. Keckler, "Compressing DMA Engine: Leveraging Activation Sparsity for Training Deep Neural Networks," in *IEEE International Symposium on High Performance Computer Architecture (HPCA)*, 2018.

[11] Y. Shi, M. Wang, S. Chen, J. Wei and Z. Wang, "Transform-Based Feature Map Compression for CNN Inference," in *Proc. IEEE International Symposium on Circuits and Systems (ISCAS)*, 2021.

[12] B. Chmiel, C. Baskin, R. Banner, E. Zheltonozhskii, Y. Yermolin, A. Karbachevsky, A. Bronstein and A. Mendelson, "Feature Map Transform Coding for Energy-Efficient CNN Inference," in *Proc. International Joint Conference on Neural Networks (IJCNN)*, pp. 1-9, 2020.

[13] F. Xiong, F. Tu, M. Shi, Y. Wang, L. Liu, S. Wei and S. Yin, "STC: Significance-aware Transform-based Codec Framework for External Memory Access Reduction," in *2020 57th ACM/IEEE Design Automation Conference (DAC)*, 2020.

[14] X. Zhang, J. Zou, X. Ming, K. He and J. Sun, "Efficient and accurate approximations of nonlinear convolutional networks," in *Proc. IEEE Conference on Computer Vision and Pattern Recognition (CVPR)*, pp. 1984-1992, 2015.

[15] O. Russakovsky, J. Deng, H. Su, J. Krause, S. Satheesh, S. Ma, Z. Huang, A. Karpathy, A. Khosla, M. Bernstein, A. C. Berg and L. Fei-Fei, "ImageNet Large Scale Visual Recognition Challenge," *International Journal of Computer Vision (IJCV)*, vol. 115, pp. 211-252, 2015.